\documentclass[a4paper]{article}

\usepackage{INTERSPEECH2019}
\usepackage{multirow}

\title{Adversarial Training for Multilingual Acoustic Modeling}
\name{Ke Hu, Hasim Sak, Hank Liao}
\address{Google Inc., USA}
\email{\{huk,hasim,hankliao\}@google.com}

\begin{document}

\maketitle
\begin{abstract}
Multilingual training has been shown to improve acoustic modeling performance by sharing and transferring knowledge in modeling different languages. Knowledge sharing is usually achieved by using common lower-level layers for different languages in a deep neural network. Recently, the domain adversarial network was proposed to reduce domain mismatch of training data and learn domain-invariant features. It is thus worth exploring whether adversarial training can further promote knowledge sharing in multilingual models. In this work, we apply the domain adversarial network to encourage the shared layers of a multilingual model to learn language-invariant features. Bidirectional Long Short-Term Memory (LSTM) recurrent neural networks (RNN) are used as building blocks. We show that shared layers learned this way contain less language identification information and lead to better performance. In an automatic speech recognition task for seven languages, the resultant acoustic model improves the word error rate (WER) of the multilingual model by 4\% relative on average, and the monolingual models by 10\%.
\end{abstract}

\section{Introduction}
\label{sec:intro}

Automatic speech recognition (ASR) has made significant progress in recent years by using deep neural networks (DNN) \cite{hinton12} for acoustic modeling. Its performance has been further improved by Long Short-Term Memory (LSTM) recurrent neural networks \cite{hochreiter1997} and large-scale training \cite{sak14lstm, sak14seq, sak15, soltau16}. One potential challenge in these modeling tasks is to obtain enough training data for different languages. To tackle the resource-scarce conditions, different approaches such as semi-supervised training, unsupervised training, and multitask learning have been proposed. Among them, multilingual training has received great interest and been shown to achieve significant improvements over monolingual models \cite{heigold13, huang13, lin09}. One benefit that multilingual training brings in acoustic modeling is to share training data from multiple languages to cover wider acoustic context.

Recently, the domain adversarial network (DAN) was designed to reduce mismatch in domains by adversarially learning domain-invariant features \cite{ganin16}. In a multitask learning framework, an adversarial head is jointly trained with a classification task to classify training samples into different domains. By reversing the gradients, the lower-level layers of the network try to minimize the domain difference and learn domain-invariant features. DAN has been applied to ASR for different purposes \cite{shinohara16,meng18,sun18,beutel17}. 

In this work, we propose to use domain adversarial training to promote the data sharing between different languages in acoustic modeling. Instead of feed-forward DNNs, we use recurrent LSTMs for acoustic modeling in a context-dependent (CD) phone state based hidden Markov model neural network system. The proposed model contains shared LSTM layers for feature extraction which aim to learn a common feature representation using data from multiple languages. The output activations of the shared layers are further used by the acoustic modeling (AM) layers for CD state classification. Each language has a separate AM head. A language identification (LID) layer tries to minimize the language classification loss, but the learned gradients are reversed to encourage shared layers to learn language-invariant features. All layers are jointly trained using data from multiple languages. We show that, when training with seven languages, feature learned indeed contains less effective information for LID compared to feature learned without adversarial layers. The acoustic modeling benefits from adversarial learning by improving the word error rate (WER) by 4\% on average compared to without adversarial training. The improvement over monolingual models is 10\%.

\section{Related work}
\label{sec:related_work}

Multilingual training has been extensively studied in ASR. A multilingual acoustic model can use a global phoneset \cite{lin09, schultz01, niesler07, vu14} or have per-language output layers \cite{heigold13,huang13,cui15}. In the latter case, Heigold et al.\ \cite{heigold13} used multitask learning to train a multilingual model for 11 Romanic languages and dialects. A three-layer DNN is used as a shared representation for feature learning, and multiple output layers are used for each language for acoustic label classification. In a similar work \cite{huang13}, Huang et al.\ trained multilingual DNN models for four languages in the context of multitask learning and also cross-lingual training. They obtained significant WER reduction in multilingual conditions and also demonstrated that knowledge learned from multiple languages can be transferred to unseen languages. Our network structure is similar to \cite{heigold13, huang13} but adversarial layers are added to reduce language difference in feature learning. More recently, multilingual (or multidialect) training has been researched in end-to-end models such as a joint attention and connectionist temporal classification (CTC) model or Listen, Attend, and Spell (LAS) \cite{toshniwal17, li17,watanabe17, kim17}. These models usually output graphemes and need to perform language identification in joint training. It is unclear whether adversarial training can improve learning language-invariant features.

Adversarial training receives considerable attention recently. It shows remarkable ability in training generative models \cite{goodfellow14} and domain-invariant training \cite{ganin16}. In ASR, DAN was used to help learning noise robust feature in multiple known and unknown noisy environments \cite{shinohara16}. In \cite{meng18}, DAN was applied to speaker invariant training and obtained superior results over regular speaker-independent ASR models. In another work, Sun et al.\ applied DAN to unsupervised training of accented speech recognizers and obtained uniform improvements in different dialects \cite{sun18}. In addition, Beutel et al.\ uses an adversarial head to discourage the model from learning gender sensitive information \cite{beutel17}.

\section{Multilingual adversarial training}
\label{sec:mla}

\begin{figure}[t]
\begin{center}
\includegraphics[height=2.5in,width=2.9in]{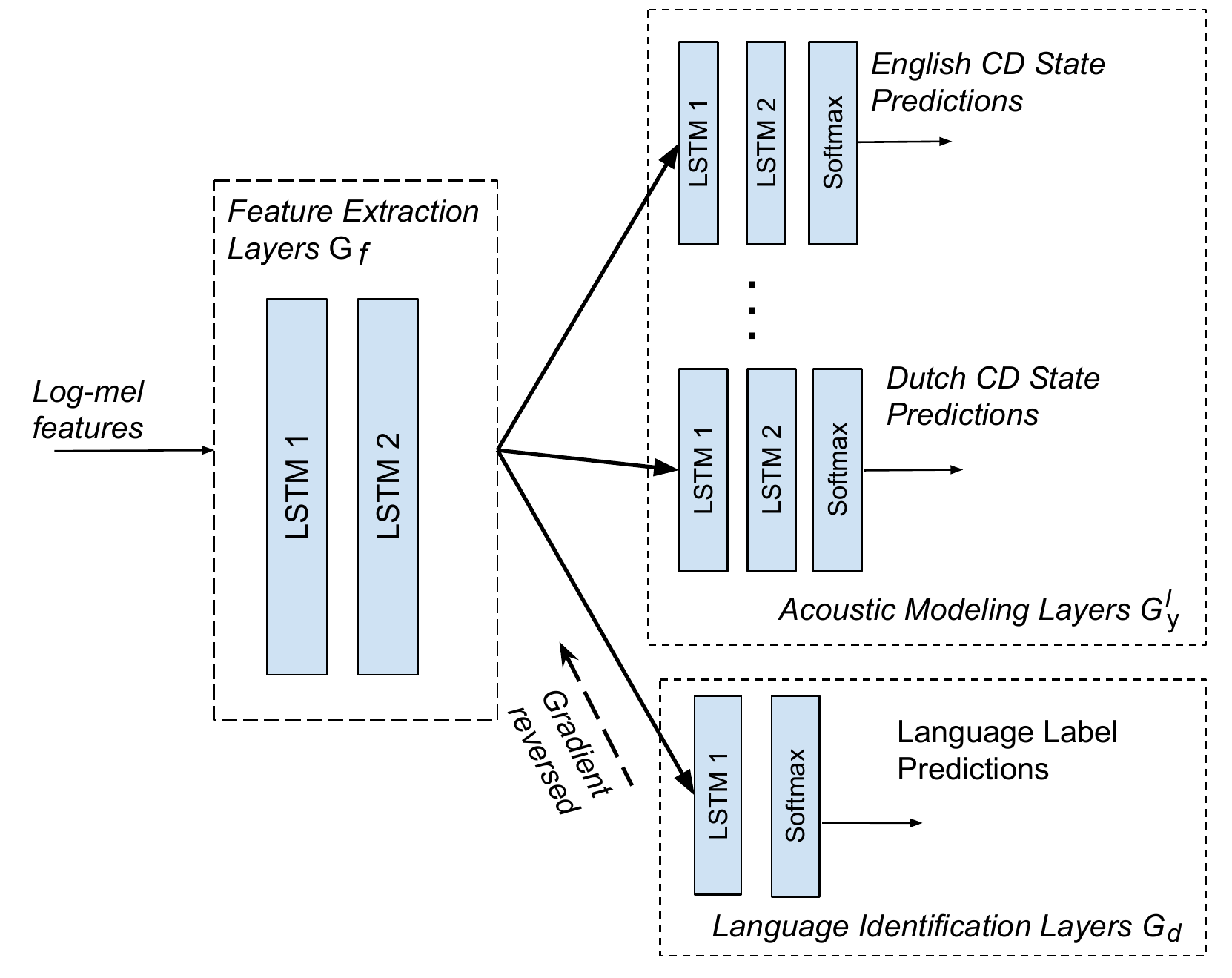}
\caption{A diagram of multilingual adversarial neural network. Gradients from the language identification layers are reversed when they are backpropagated (dashed line).}
\label{fig:mla}
\end{center}
\end{figure}

Given supervised training samples $\{x^l_i, y^l_i\}$, where $x^l_i$ is the input and $y^l_i$ the acoustic label, our goal is to learn a multilingual model to estimate acoustic labels for every language. Here, the subscript $i=1,...,N$ denotes the number of training samples, and superscript $l=1,...,L$ the number of languages.  As shown in Fig. \ref{fig:mla}, different languages share the same lower feature extraction layers, however at higher levels each language has their own AM layers. Therefore, we do not merge the phone sets of different languages. We denote the feature extraction layers as $G_f$ with parameters $\theta_f$, and the AM layers as $G^l_y$ with parameters $\theta^l_y$, where $l$ indexes the language. Since each language is considered as a different domain, we attach LID layers to the feature extraction layers to predict frame-level language labels. The LID layers are denoted as $G_d$ with parameters $\theta_d$. 

According to \cite{ganin16}, the overall loss function for DAN is

\begin{align}
    & \mathcal{L}(\theta_f, \{\theta^l_y\}_{l=1,...,L}, \theta_d) = \notag \\
    & \mathcal{L}_{AM}(\theta_f, \{\theta^l_y\}_{l=1,...,L}) - \lambda \mathcal{L}_{LID}(\theta_f, \theta_d)
\end{align}
where $\mathcal{L}_{AM}(\cdot)$ is the average cross-entropy (CE) loss from CD state classification for all languages, and $\mathcal{L}_{LID}(\cdot)$ the frame-level CE loss for LID, and $\lambda > 0$ is the adversarial weight which will be multiplied to the reversed gradients when it is backpropagated from the LID layer. Note that 
when minimizing $\mathcal{L}_{AM}$, $\theta^l_y$ is always adjusted to minimize the AM loss, and the error signals are backpropagated to optimize $\theta_f$. On the other hand, $G_f$ and $G_d$ are jointly trained with an adversarial loss $\mathcal{L}_{LID}$, where $\theta_f$ is adjusted to maximize the loss, and $\theta_d$ is adjusted to minimize the loss. The two play a minimax game where $G_d$ tries to discriminate inputs from different languages using features generated by $G_f$, while $G_f$ tries to generate features to confuse $G_d$ to not make the right domain classification decision. For gradient reversal, we follow the implementation in \cite{ganin16} and keep the propagation unchanged in the forward path, and multiply the gradient by $-\lambda$ during backpropagation. We update the parameters by back propagation using stochastic gradient descent:

\begin{align}
    & \theta_f \leftarrow \theta_f - \alpha(\frac{\partial \mathcal{L}_{AM}}{\partial \theta_f} - \lambda \frac{\partial \mathcal{L}_{LID}}{\partial \theta_f}) \\
    & \theta^l_y \leftarrow \theta^l_y - \alpha\frac{\partial \mathcal{L}_{AM}}{\partial \theta^l_y} \\
    & \theta_d \leftarrow \theta_d - \alpha\lambda\frac{\partial \mathcal{L}_{LID}}{\partial \theta_d}
\end{align}
where $\alpha$ is the learning rate.


\section{Evaluation and Comparison}
\label{sec:exp}

\subsection{Data and Training Setup}
We train the models based on seven languages including English, Spanish, French, Portuguese, Italian, Russian and Dutch. The amount of training data for different languages are given in Table \ref{tab:1}. Both of our training and test data contain publicly uploaded videos from different categories similar to \cite{liao13, soltau16}. These languages come from different groups including Romance, Germanic, and Slavic languages. The training data are manually segmented and transcribed yielding high-quality transcripts for training. To obtain alignments for CE training, the data are force-aligned by using acoustic model trained by supervised training data (ranging from 71 to 363 hours) for each language. The test data contain at least 5 hours of data for each language. Note that for Dutch, we purposely use a small amount of training data, i.e. 13.5 hr, to simulate a low-resource condition. We want to analyze how such a low-resource language can benefit from multilingual and adversarial training. The inputs to the models are 80-dimensional log-mel filterbank energy features based on a 25-ms frames width a 10-ms frame rate. For each language, a new state inventory consisting of 1600 CD states is generated.

Our baseline models were monolingual models, each consisting of 4 layers of bidirectional LSTMs with 200 memory cells each direction in each layer, and trained only on the data available for a given language.  A single multilingual model, without the LID layers \cite{heigold13}, was trained on the data available for the monolingual models but as a single, randomly shuffled data set; multiple, per language, softmax ``heads" follow the shared 4-layer LSTM stack trained only on the matching language data.

For multilingual adversarial training, we use 2 bidirectional LSTM layers each with 200 memory units in each direction as the shared layers, 2 bidirectional LSTM layers of 200 cells each direction as AM layers for each language, and 1 bidirectional LSTM layer with 128 cells each direction for LID. The AM layers of a particular language will be updated only when the corresponding training data is presented. In training of the adversarial layers, we removed silence frames which does not contain any language information. An adversarial weight ($\lambda$) of 1 is multiplied with the reversed gradients. We have also tried other weights and that does not significantly change the performance. Our training is done in parallel on CPUs using asynchronous stochastic gradient descent (ASGD) training described in \cite{sak14lstm,sak15,sak14seq,heigold13}. We choose the best model to be the one with highest frame accuracy on the development set. The frame accuracy is calculated by uniformly sampling utterances from different languages. For decoding, a standard 5-gram language model is trained.

\subsection{Multilingual Adversarial Training}

\begin{table*}[t]
  \centering
    \begin{tabular}{ |c|c||c|c|c| }
      \hline
      \multirow{2}{*}{Language} & \multirow{2}{*}{Training Data (hr)} & \multicolumn{3}{|c|}{WER (\%) of different models} \\ \cline{3-5}
      & & Monolingual & Multilingual & Adversarial \\ 
      \hline\hline
      English & 363 & 38.6 & 33.0 & 36.9 \\ \hline
      Spanish & 141 & 27.1 & 26.2 & 24.2 \\ \hline
      French & 141 & 42.7 & 45.4 & 41.3 \\ \hline
      Portuguese & 118 & 34.4 & 33.7 & 31.4 \\ \hline
      Italian & 117 & 41.0 & 40.4 & 36.9 \\ \hline
      Russian & 71 & 53.3 & 51.0 & 47.7 \\ \hline
      Dutch & 13.5 & 63.1 & 52.5 & 51.7 \\ 
      \hline\hline
      \multicolumn{2}{|c||}{Average} & 42.9 & 40.3 (-6\%) & 38.6 (-10\%) \\
      \hline
    \end{tabular}
    \caption{Comparison of the WERs across monolingual, multilingual and multilingual adversarial models for each language. The average relative WER reductions w.r.t. monolingual models are shown in parentheses.}
    \label{tab:1}
\end{table*}

Table \ref{tab:1} shows WERs of monolingual, multilingual and multilingual adversarial models. For monolingual models, we note that the relative high WERs for Russian and Dutch are due to their insufficient training data. Multilingual training improves WER for almost all languages except French. The improvement for Dutch is the most, i.e. a relative 17\%. The biggest improvement for Dutch confirms that multilingual training benefits significantly for low-resource languages, as observed in \cite{huang13} and \cite{heigold13}. The degradation for French indicates that multilingual data introduces noise for its feature representation. However, we will show that adversarial training improves in this condition. Overall, the multilingual model performs better than monolingual models by 6\% in terms of WER.

Based on the multilingual model, we observe in Table \ref{tab:1}  that almost all languages obtain significant WER reductions after adversarial training. The WER reduction ranges from 1.5\% to 9\%. The improvement for Dutch is reduced compared to the monolingual-to-multilingual gain since it has obtained significant gain from multilingual training. The gains brought by adversarial training come from enforcing the shared layers to learn language-invariant features, and it seems that most languages benefit from this enhanced knowledge sharing. The degradation of English may be due to that it already gained large improvement from monolingual to multilingual training, and learning language-invariant features reduced the benefits. The relatively more training data from English may be responsible for the regression, and one future experiment is to create more balanced training sets across languages, for example, using data mining technique in \cite{soltau16}. In summary, the multilingual adversarial model performs uniformly and significantly better than monolingual models for all languages, and the average WER reduction is around 10\%. We will further analyze the outputs from the shared layers to confirm whether language specific information is attenuated in Sect. \ref{sec:lid}. On average, adversarial training brings an additional relative 4\% WER reduction over multilingual training.

In addition to using a 2-layer bidirectional LSTM as feature extractor, we have also tried varying the complexity of the shared layers, i.e. using a one- or three-layer LSTM. In these scenarios, the LID layers are still attached to output of the shared layers, and the total number of feature extractor and AM layers remain constant to make sure models have the same capacity. We find out that the acoustic model performs worse when using a single layer feature extractor, the average WER increases to 43.6\%, i.e. around 13\% worse than a two-layer feature extractor. We attribute this to that it is a very early layer in the deep structure and enforcing it to learn language invariant features hurts the effectiveness of acoustic modeling. On the other hand, when using a three-layer feature extractor, our training converges very slowly and eventually to substantially higher CE loss (almost doubled). This is probably because the feature extractor is too strong compared to the adversarial layers, and forcing learning language-invariant features results in loss of necessary phonetic information for CD state classification.

\subsection{Language Identification Performance}
\label{sec:lid}

To confirm the effectiveness of adversarial training, it is interesting to see how much language identification information still remains in the outputs from feature extractor. In order to quantify that, we trained a linear classifier using the activations from the feature extractor for LID and the frame-level CE loss. Here the feature extractor is fixed during training to maintain the feature representation learned during earlier training. We measure the average frame-level LID accuracies on the dev sets and found that adversarial training produces 21.6\% relative lower average frame accuracy than multilingual training (48.8\% vs 62.2\%). This shows that adversarial training is effective in removing LID features from the feature extractor. On the other hand, we note that not all LID information is lost, and this is because phonetic information preserved by AM layers is helpful for LID.

\subsection{Per-Language Fine Tuning}

In addition to adversarial training, we are curious to see whether we can further improve the performance of each language by fine tuning the multilingual or multilingual adversarial model using per-language training data, similar to \cite{huang13}. We thus train a monolingual model for each language, i.e. 4-layer bidirectional LSTM with 200 cells each direction at a layer, by initializing its parameters from the shared and corresponding AM layers of the multilingual (or adversarial) model. The training then continues for all parameters until the frame accuracy does not improve. The same learning rate is used as in monolingual training. 


We find that fine-tuning for multilingual models benefits nearly all languages. The WER improvement is around 4.5\% on average (Table \ref{tab:2}). This means most languages benefit from further biasing the feature representations towards its own language while not losing shared knowledge from other languages. We only see degradation in English and it is because fine tuning results in loss of the shared knowledge due to its relatively large amount of data. We confirmed this by fine tuning only the AM layers for English and achieved a comparable WER (33.8\%) as the multilingual model.

On the other hand, we also fine tune the multilingual adversarial models for each language. It is interesting to see that most languages perform worse, which leads to 2.8\% higher average WER (Table \ref{tab:3}). This may mean that the language-invariant feature representation is sensitive to per-language fine tuning and was partly lost. One future experiment would be to fine-tune different layers and compare WER changes. On the other hand, since the shared layers are initialized from the multilingual adversarial model, the fine-tuned models still perform significantly better than monolingual models.

\begin{table}
\centering
\begin{tabular}{ |c||c|c| }
    \hline
    \multirow{2}{*}{Language} & \multicolumn{2}{|c|}{WER (\%)} \\ \cline{2-3}
    & Multilingual & +Fine Tuning \\ 
    \hline\hline
    English & 33.0 & 36.0 \\ \hline
    Spanish & 26.2 & 24.2 \\ \hline
    French & 45.4 & 40.8 \\ \hline
    Portuguese & 33.7 & 31.3 \\ \hline
    Italian & 40.4 & 37.1 \\ \hline
    Russian & 51.0 & 49.1 \\ \hline
    Dutch & 52.5 & 51.5 \\ \hline
    \hline
    Average & 40.3 & 38.5 (-4.5\%) \\ \hline
\end{tabular}
\caption{WERs (\%) of multilingual models before and after fine-tuning. }
\label{tab:2}
\end{table}

\begin{table}
\centering
\begin{tabular}{ |c||c|c| }
    \hline
    \multirow{2}{*}{Language} & \multicolumn{2}{|c|}{WER (\%)} \\ \cline{2-3}
    & Adversarial & +Fine Tuning \\ 
    \hline\hline
    English & 36.9 & 36.4 \\ \hline
    Spanish & 24.2 & 24.8 \\ \hline
    French & 41.3 & 41.4 \\ \hline
    Portuguese & 31.4 & 32.3 \\ \hline
    Italian & 36.9 & 39.9 \\ \hline
    Russian & 47.7 & 48.9 \\ \hline
    Dutch & 51.7 & 54.3 \\ \hline
    \hline
    Average & 38.6 & 39.7 (+2.8\%) \\ \hline
\end{tabular}
\caption{WERs (\%) of multilingual adversarial models before and after fine-tuning. }
\label{tab:3}
\end{table}

\section{Concluding Remarks}
\label{sec:conclude}
In this work we explored applying domain adversarial networks to multilingual training. The results show that features learned from the shared layers contain less language identification information, and encouraging the learning of language-invariant features improves the performance of most languages. Adversarial learning was applied to a 4-layer bidirectional LSTMs, and it matters that to which hidden layer we apply the adversarial learning. On average, adversarial learning improves the WER of the multilingual model by 4\% and the monolingual model by 10\%.  Fine tuning significantly improves the multilingual model but leads to regression for adversarial models, which indicates language-invariant representation is sensitive to per-language data. One interesting future work would be to fine-tune different layers of the adversarial model to analyze the regressions. Also, in this work the multilingual model consists of languages from different language groups and there is no unseen language in testing. It would be interesting to see how adversarial training performs for languages in the same group, or how it generalizes to unseen languages using Babel data in \cite{cui15}.

\section{Acknowledgements}
The authors want to thank Hagen Soltau, Shankar Kumar, and anonymous reviewers for their useful comments and discussions.

\bibliographystyle{IEEEtran}

\bibliography{mybib}


\end{document}